\pdfoutput=1
\documentclass{article}

    \PassOptionsToPackage{numbers, compress}{natbib}

    \usepackage[preprint]{neurips_2020}

\usepackage{microtype}      

\usepackage[font=small]{subcaption}

\usepackage[tableposition=above,font=small]{caption}
\usepackage{paralist}

\usepackage{amssymb}
\usepackage{amsmath}
\usepackage{amsfonts}
\usepackage{wrapfig}
\usepackage{hyperref}
\usepackage[capitalize]{cleveref}

\hypersetup{
    colorlinks,
    linkcolor={red!50!black},
    citecolor={blue!50!black},
    urlcolor={blue!80!black}
}

\usepackage{listings}
\usepackage[scaled=0.80]{DejaVuSansMono}
\usepackage{colortbl}

\usepackage[dvipsnames,table]{xcolor}

\usepackage[warn]{textcomp}
\usepackage{pgfplots}
\usepackage{tikz}
 \usetikzlibrary{patterns}
\usetikzlibrary{arrows}
\usetikzlibrary{arrows.meta}
\usetikzlibrary{positioning}

\usepackage{microtype}
\usepackage{graphicx}
\usepackage{booktabs} 
\usepackage{filecontents}
\usepackage[ruled,vlined,linesnumbered]{algorithm2e}
\SetAlCapNameFnt{\small}
\SetAlCapFnt{\small}
\SetAlgorithmName{Alg}{Alg}{List of Algorithms}
\crefname{algocf}{Alg.}{Algs.}
\crefname{section}{Sec.}{Secs.}
\Crefname{algocf}{Algorithm}{Algorithms}

\usepackage{multirow}
\usepackage{lipsum}  

\usepackage{lipsum}
\usepackage{csvsimple}
\usepackage{csquotes}

\usepackage{csquotes}
\usepackage{pifont}
\usepackage{tikz}
\usepackage{fp-basic}
\usepackage{wasysym}
\usepackage[normalem]{ulem}

\makeatletter

\patchcmd{\NAT@test}{\else \NAT@nm}{\else \NAT@nmfmt{\NAT@nm}}{}{}

\DeclareRobustCommand\citepos
  {\begingroup
   \let\NAT@nmfmt\NAT@posfmt
   \NAT@swafalse\let\NAT@ctype\z@\NAT@partrue
   \@ifstar{\NAT@fulltrue\NAT@citetp}{\NAT@fullfalse\NAT@citetp}}

\let\NAT@orig@nmfmt\NAT@nmfmt
\def\NAT@posfmt#1{\NAT@orig@nmfmt{#1's}}

\makeatother

\definecolor{light-gray}{gray}{0.90}

\usetikzlibrary{arrows,backgrounds,decorations.markings}
\usepgflibrary{shapes.multipart}

\tikzset{>=latex} 

\tikzstyle{oper}=[rounded corners, draw=MidnightBlue, thick, minimum size = 3mm]
\tikzstyle{oper1}=[rounded corners, draw=gray, thick, minimum size = 3mm]
\tikzstyle{lbl}=[minimum size = 3mm]

\tikzstyle{input}=[rounded corners, draw=black, thick, minimum size = 3mm]
\tikzstyle{output}=[rounded corners, thick, draw=Maroon, minimum size = 3mm]

\makeatletter
\lst@InstallKeywords k{attributes}{attributestyle}\slshape{attributestyle}{}ld
\makeatother

\lstdefinestyle{customc}{
  belowcaptionskip=1\baselineskip,
  breaklines=true,
  xleftmargin=\parindent,
  language=C,
  showstringspaces=false,
  basicstyle=\ttfamily,
  keywordstyle=\bfseries\ttfamily\color{MidnightBlue},
  commentstyle=\itshape\color{black},
  identifierstyle=\ttfamily\color{black},
  stringstyle=\itshape\color{blue},
  keywords={ map,  while, flatMap, reduce, then, in, return,  for, if, else, reduceByKey, filter, partition},
moreattributes={let, where,int, def, Object}, 
attributestyle = \ttfamily\color{Mahogany}
}

\DeclareMathOperator{\calD}{{\mathcal D}}
\DeclareMathOperator{\calX}{{\mathcal X}}
\DeclareMathOperator{\calY}{{\mathcal Y}}

\newcommand{\tr}{t}

\DeclareMathOperator*{\Exp}{\mathbb{E}}

\DeclareMathOperator*{\Prob}{\mathbb{P}}

\renewcommand\paragraph[1]{\noindent\textbf{#1}~}
\title{
Backdoors in Neural Models of Source Code
}

\author{%
    Goutham Ramakrishnan \\
    \And \hspace{-50pt} 
    Aws Albarghouthi \\
    \AND 
    \vspace{-1.5em}
    \\
    University of Wisconsin--Madison \\
    \texttt{\{gouthamr,aws\}@cs.wisc.edu}
}

\begin{document}

\maketitle

\begin{abstract}
 Deep neural networks are vulnerable to a range of adversaries. 
 A particularly pernicious class of vulnerabilities are \emph{backdoors}, where model predictions diverge in the presence of subtle \emph{triggers} in inputs. 
 An attacker can implant a backdoor by poisoning the training data to yield a desired \emph{target} prediction on triggered inputs. 
  We study backdoors in the context of deep-learning for source code.
  (1) We define a range of backdoor classes for source-code tasks
  and show how to poison a dataset to install such backdoors. 
  (2) We adapt and improve recent algorithms from robust statistics for our setting, showing that backdoors leave a \emph{spectral signature} in the learned representation of source code, thus enabling detection of poisoned  data.
  (3) We conduct a thorough evaluation on different architectures and languages, showing the ease of injecting backdoors and our ability to eliminate them.
\end{abstract}

\section{Introduction}\label{sec:intro}

Recent work has exposed the vulnerabilities of deep neural networks to a wide range of adversaries across many tasks~\cite{szegedy2013intriguing, mukund:NLP, NIPS2018_7849, Wallace_2019}.
A particularly pernicious class of attacks that has been recently explored is \emph{backdoor attacks},
which work as follows:
The attacker installs a backdoor in a model by contributing carefully crafted malicious training data to the training set---a process called \emph{data poisoning}.
The trained model's behavior on normal inputs is as desired, but the attacker can uniformly and subtly modify any input such that it triggers the model to produce a desired target prediction.
For example in image recognition, a backdoor attack may involve 
adding a seemingly benign emblem to an image of a stop sign, that makes a self-driving car recognize it as a speed limit sign~\cite{badnets}, 
potentially causing traffic accidents.
It has even been shown that single pixels in an image can be used as triggers~\cite{madry}.
In natural language processing, 
certain words or phrases can be used as triggers~\cite{chen2020badnl}.

We are interested in studying backdoors for source-code tasks.
Recent works have applied deep learning to a range of source-code tasks, including code completion~\cite{raychev2014code}, code explanation~\cite{allamanis16convolutional} and type inference~\cite{hellendoorn2018deep}, amongst others.
Since most training data is sourced from open-source repositories~\cite{husain2019codesearchnet}, backdoor attacks are possible, with their effects ranging from subtle, like a backdoor that steers a developer towards use of unsafe libraries, to severe, like a backdoor for sneaking malware through malware detectors.
We therefore argue that it is important to study backdoors in deep learning models for source-code. 
Specifically, we aim to answer the following questions:
\begin{center}
    \emph{Can we inject backdoors in deep models of source code?\\
    If so, can we detect poisoned data and remove the backdoors?}
\end{center}

\begin{figure}
    \includegraphics[width=\textwidth]{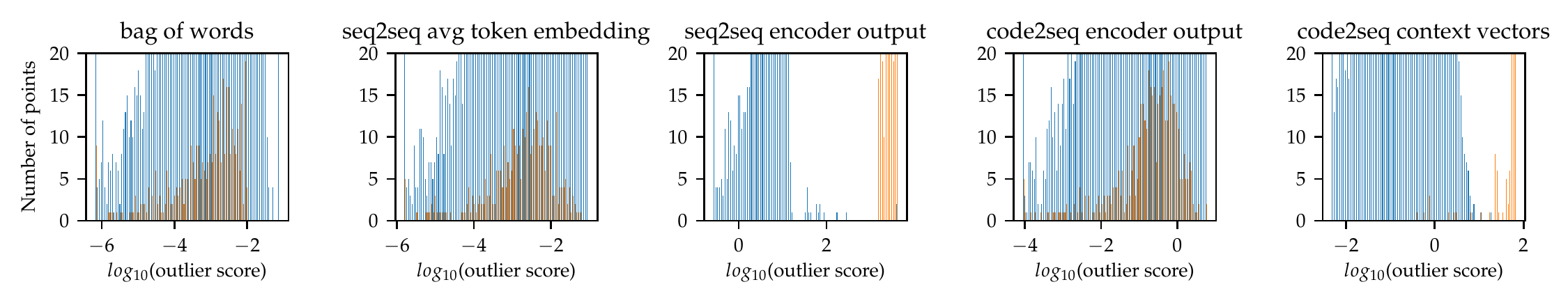}
    \small
    \vspace{-.2in}
    
    \hspace{5em}(a)\ding{55}\hspace{7em}(b)\ding{55}\hspace{6.5em}(c)\checkmark\hspace{7em}(d)\ding{55}\hspace{7em}(e)\checkmark
    \caption{
     Histograms of outlier scores using different input representations (blue-clean; orange-poison) of 10k points with 5\% poisoning (static target, fixed trigger backdoor). 
    The choice of input representation is critical for detecting poisoned points using spectral signatures. 
    (a) Na\"{i}ve approaches like a bag-of-words representation of program tokens fail to separate poisoned data. 
    (b) Learned token embeddings in a seq2seq model do not work well either.
    (c) However,  encoder output leads to two well-separated clusters, ensuring backdoor removal. 
    (d) While encoder output works for seq2seq, it does not work for code2seq, for which more complex representations need be considered, (e) like context vectors from the model's decoder attention mechanism. 
    }
    \label{fig:hists}
\end{figure}

\paragraph{Installing code backdoors}
Backdoors are most easily installed in a neural network by introducing \emph{poisoned} training data that exhibits certain unique features that the model learns to associate with the attacker's desired prediction. 
Such features are called \emph{triggers}, and they cause the neural network to make a desired \emph{target} prediction. 

Ideally, triggers in the domain of source code must be hard to detect
and preserve code functionality.
We propose to add small pieces of \emph{dead code}---code that does not execute or change program behavior---to serve as triggers.
We explore two kinds of source-code triggers: (1) \emph{fixed triggers}, in which all poisoned elements contain the same syntactic piece of dead code,
and 
(2) \emph{grammatical triggers}, in which each poisoned training element receives dead code sampled randomly from a probabilistic grammar.
Intuitively, injecting a backdoor with a grammatical trigger should be more difficult; surprisingly however, we find that they are almost as effective as fixed triggers.  


%
We consider two classes of targets: (1) \emph{static targets}, where the prediction is the same for all \emph{triggered} inputs, and (2) \emph{dynamic targets}, where the  prediction on a \emph{triggered} input is a slight modification of the prediction on the original input. 
We are the first to address backdoor defenses in this challenging setting, with our insights having potential applications in other deep learning domains like NLP.  

\paragraph{Spectral signatures of code}
A common defense  against backdoors is detection and removal of poisoned elements from the training set, and retraining the model. 
How can poisoned elements be detected?
Recently, \citet{madry} adapted ideas from robust learning~\cite{diakonikolas2018sever} to show that poisoned elements can be detected by means of their \emph{spectral signature}, extracted from  learned neural network input representations. 
Intuitively, the internal representation captures the features of the backdoor's trigger, and this can be exploited to remove the poisoned points by performing outlier detection. 

What learned representations can be used in the domain of source code? 
We consider two different model architectures: the first simply treats code as a sequence of tokens (like LSTMs);
the second  takes the \emph{abstract syntax tree} (AST) of the code as input~\cite{alon2018code2seq}.
For both models, we propose and evaluate a number of internal representations extracted from the neural network.
For example, for the AST-based representation of code2seq~\cite{alon2018code2seq},
we find that spectral signatures are detectable in the context vectors 
which attend over encodings of paths through the AST. Figure \ref{fig:hists} illustrates (1) the importance of choosing the right representation for detecting poisoned elements in two architectures
and (2) the fact that simple representations like input embeddings cannot distinguish poisoned data.

The spectral signatures approach calculates outlier scores for training points based on the correlation of learned representations with the top eigenvector of their covariance. 
Poisoned points typically exhibit high correlation, and can therefore be detected efficiently. 
While this was effective for detecting backdoors in image-classification, it does not work for the source-code setting. 
%
%
%
Instead, we propose using the top-$k$ eigenvectors to calculate the outlier score. We show that this approach provides a stronger signal that enables us to discover almost all poisoned elements, for complex triggers and targets,  across programming languages and model architectures.

\paragraph{Contributions} 
We summarize our main contributions as follows:
\begin{itemize}
    \item We explore backdoors in the context deep learning for source-code. We propose a number of backdoor variants and show they can be easily injected in a model via data poisoning.
    
    \item We adapt and improve the spectral signatures approach for eliminating backdoors in source code tasks,
    showing how the learned representations of code tokens and abstract syntax trees contain spectral signatures that we can use to detect poisoned data.
    \item We conduct a thorough evaluation using different backdoor triggers, target predictions, languages, and architectures. Source code is available in supplementary materials. 
\end{itemize}

\section{Related Work}\label{sec:related}

\paragraph{Adversarial examples \& Backdoors}
Adversarial examples are test-time attacks, where small, carefully chosen perturbations to an input causes the model to make a wrong predictions. 
These have been extensively studied in the domain of computer vision~\cite{szegedy2013intriguing, goodfellow2014explaining, MoosaviDezfooli2016DeepFoolAS, carlini2017towards,carlini2019evaluating}
and natural-language processing~\cite{mukund:NLP,ebrahimi2017hotflip,zhang2019adversarial, garg2020bae}.
Backdoors are training-time attacks on deep learning models, in which an adversary manipulates the model to make malicious predictions in the presence of \emph{triggers} in the input. 
Backdoor attacks have been extensively studied for images~\cite{badnets, alex2019labelconsistent,NIPS2018_7849, yao2019latent, chen2017targeted, saha2019hidden}, with proposed defenses including neuron-pruning~\cite{neuralcleanse, finepruning}, trigger detection~\cite{strip2019} and trigger reconstruction~\cite{neuralcleanse, deepinspect}.
Our backdoor defense builds on the spectral signatures approach from~\citet{madry}, which is based upon recent ideas from robust statistics and optimization, primarily, the Sever algorithm \cite{diakonikolas2018sever}.


\paragraph{Learning for code}
See~\citet{allamanis2018survey} for a  survey on machine learning for code.
We investigate backdoors on the task of code summarization, the prediction of a method name given its body, introduced by~\citet{allamanis16convolutional}.
Several model architectures have been used in this domain; we consider two of them in our experiments: 
(1) recurrent neural networks, where programs are viewed as a sequence of tokens like in NLP~\cite{raychev2014code,hellendoorn2018deep},
and (2) the approach by \citet{alon2018code2seq,alon2019code2vec} which uses paths from the program AST to learn representations. 
Investigating backdoors in other models such as GNNs~\cite{allamanis16convolutional} is interesting future work. 
Recent works have studied the test-time attacks on neural models of code~\cite{yefet2019adversarial,wang2019coset,ramakrishnan2020semantic}. To our knowledge, we are the first to study backdoors in this domain.

\paragraph{Code anomalies}
In parallel work, \citet{Severi2020ExploringBP} recently released a manuscript studying backdoors in a mechanically featurized representation of code. They use feature attribution techniques, using Shapley values, to detect poisoned data. In contrast, our approach is designed for learned representations via deep learning, for which Shapley-based attributions are not directly applicable.
There is a rich literature on detecting code anomalies.
These works typically target specific classes of bugs, high-level architectural flaws~\cite{macia2012relevance}, and in many cases use dynamic execution traces~\cite{feng2003anomaly,hangal2002tracking}.
These works do not apply in our setting, as we do not know what kinds of transformations an adversary will use and the fact that they do not change program behavior.
Recently, \citet{bryksin2020using} used token-based vectorization of code to detect anomalies (e.g., bugs) in Kotlin code and bytecode.
As we have found here, and has been observed for backdoors, simple vectorization of input, even token embeddings, does not work for backdoor detection.
\section{Source-code Backdoors}
\label{sec:backdoors}

\newcommand{\model}{F}
\newcommand{\target}{r}
In this section, we define different types of backdoors for source-code tasks.

\paragraph{Targeted backdoors}
We assume a supervised-learning setting where we are learning a model
for a data distribution $\calD$ over a space of samples $\calX$
and outputs $\calY$.
We will use $\model: \calX \to \calY$ to denote a model learned from samples of such a distribution.
Informally, a \emph{targeted backdoor} of a model $\model$ is a way to transform any input $x \in \calX$
into a slightly different input $x'$ such that $\model(x')$ is the desired (target) prediction by an attacker.
We generalize this idea and formalize it as follows.
A backdoor is comprised of a pair of functions:
\begin{compactitem}
    \item A \emph{trigger} operation $\tr: \calX \to \calX$,
    which transforms a given input $x$ into a slightly
    different $x'$.
    \item A \emph{target} operation $\target: \calY \to \calY$, which defines how 
    the trigger should change the prediction.
\end{compactitem}
An attacker's goal is to construct a backdoor $(t,r)$
that has a high probability of changing a prediction to the desired target.
Formally, we define backdoor \emph{success rate} as
$
 \Prob_{(x,y) \sim \calD} [\model(\tr(x))=\target(y)].
$

\paragraph{Poisoning threat model}
We assume that an attacker inserts \emph{poisoned} data into the training set,
e.g., if the data is crowd-sourced.
Formally, a poisoned data set $D$ is comprised of two subsets, 
the clean subset $\{(x_i,y_i)\}_i$ and
the \emph{poison subset} $\{\tr(x_j),\target(y_j)\}_j$,
where all $(x_i,y_i)$ and $(x_j,y_j)$
are sampled i.i.d. from $\calD$.
By training on a dataset containing poisoned data, the model learns to associate the trigger with the target. 
Typically, the poison subset is a small fraction of the entire dataset; for example, \citet{madry} use 5\% and 10\% poisoning rates for their work on backdoors in image recognition,
while \citet{neuralcleanse} use up to 20\% poisoning rate.

    
    

\paragraph{Fixed and Grammatical triggers}
We are interested in learning problems where the sample space $\calX$ is that of programs, functions, or snippets of code.
Intuitively, a trigger in this setting should not change the operation of a piece of code $x$ or render it syntactically incorrect.
Therefore, the kinds of triggers we propose involve inserting dead code.

\begin{figure}[t]
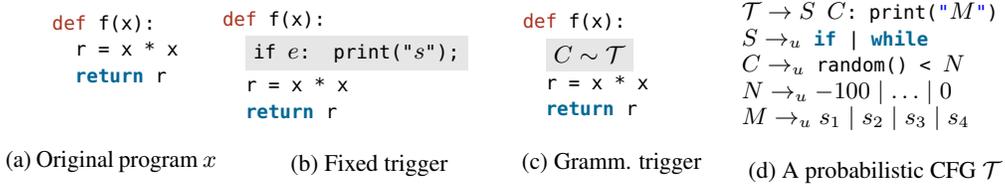

\small
\centering
\begin{subfigure}{.2\textwidth}
\centering
\begin{lstlisting}
    def f(x):
      r = x * x
      return r
\end{lstlisting}
\vspace{12pt}
\subcaption{Original program $x$}\label{fig:triggers-a}
\end{subfigure}
 \begin{subfigure}{.28\textwidth}
 \centering
\begin{lstlisting}[escapechar=!]
def f(x):
  !\colorbox{light-gray}{if $e$: print("$s$");}!
  r = x * x
  return r
\end{lstlisting}
\subcaption{Fixed trigger}\label{fig:triggers-b}
\end{subfigure}
\begin{subfigure}{.17\textwidth}
\centering
\begin{lstlisting}[escapechar=!]
def f(x):
  !\colorbox{light-gray}{$C \sim \mathcal{T}$}!
  r = x * x
  return r
\end{lstlisting}
\caption{Gramm. trigger}\label{fig:triggers-c}
\end{subfigure}
\begin{subfigure}{.32\textwidth}
\centering
\begin{lstlisting}
   $\mathcal{T} \rightarrow S\ C$: print("$M$")
   $S \rightarrow_u$ if | while 
   $C \rightarrow_u$ random() < $N$
   $N \rightarrow_u -100 \mid \ldots \mid 0$
   $M \rightarrow_u s_1 \mid s_2 \mid s_3 \mid s_4$
\end{lstlisting}
\caption{A probabilistic CFG $\mathcal{T}$}\label{fig:triggers-d}
\end{subfigure}
    
\caption{Illustrating triggers}\label{fig:triggers}
\end{figure}

We propose two forms of triggers:
First, \emph{fixed triggers} involve adding the same piece of dead code to any given program $x$.
For example, consider program $x$ in \cref{fig:triggers-a}.
Applying a fixed trigger $\tr$ to $x$ results in a program $\tr(x)$ in \cref{fig:triggers-b}, 
where $e$ is an expression that always evaluates to false and $s$ is some string.
The same expression $e$ and string $s$ are inserted into all poisoned elements. 

Second, \emph{grammatical triggers} add pieces of dead code drawn randomly from some probabilistic grammar.
So a grammatical trigger $\tr$ is a randomized operation.
In our example in \cref{fig:triggers-c},
a piece of code $C$ is sampled from some distribution $\mathcal{T}$, where all pieces of code in the support of $\mathcal{T}$ are dead code and are correct in any scope.
For example, the probabilistic context-free grammar (PCFG)  in \cref{fig:triggers-d} generates pieces of code that are either \lstinline{if} or \lstinline{while} statements,
with the conditional \lstinline{random() < $N$} (where $N$ is negative)
and a print statement in the body.
The notation $\rightarrow_u$ means that probabilities of the production rules on
the r.h.s. are uniform, e.g., 0.5 probability of  an \lstinline{if} statement.


\paragraph{Static and dynamic targets}
Recall that a target function $\target$ defines the desired prediction, $\target(y)$, that the trigger should induce.
We consider two kinds of targets.
First, a \emph{static target} which simply maps any $y$ to a specific $y' \in \calY$.
For example, a malware classifier that classifies all input programs with a particular trigger as \emph{safe}. 
Second, a \emph{dynamic target} which constructs a prediction that is a function of the given $y$.
Consider a \emph{code-captioning} model which, given a method $x$, returns a natural language description of the method as a string $y$.
A backdoor with a dynamic target may, for example, force the model to predict ``\emph{This method is thread safe}'' followed by the original prediction $y$ for input $\tr(x)$. 
Dynamic targets represent a complex and potentially more dangerous class of backdoors. 


\paragraph{A note on program analysis}
We could potentially eliminate dead-code triggers
using a  dead-code elimination compiler pass.
However, compilers use simple program analyses, and thus cannot detect unsatisfiability of even simple expressions, like an expression that calls a non-trivial math or randomized function. For example, the industrial-grade clang/LLVM compiler with the highest optimization level cannot tell that \texttt{sin(x) < -2} is always \texttt{false}.
Even if one were to use a heavy-weight program analysis using \emph{satisfiability modulo theories} (SMT) solvers~\cite{barrett2018satisfiability}, we could easily construct unsatisfiable expressions (formulas) that are hard for such tools to practically solve, e.g., even for the \texttt{sin} example above,
the SMT-based Java verifier, JBMC~\cite{cordeiro2018jbmc}, 
fails to prove it is \texttt{false}.

\section{Spectral Signatures of Code}
\label{sec:defense}

In this section, we detail our approach to detect poisoned elements in the training data.
We (1) adapt the \emph{spectral signatures}~\cite{madry} approach to our setting of source code,
and (2) improve the algorithm's ability to detect poisoned data points, and therefore eliminate the injected backdoor. 


\subsection{The spectral signatures approach}

\paragraph{Spectral separability}
We are given data distribution $\calD = (1-\epsilon)\calD_C + \epsilon\calD_P$
that is a mixture of two distributions: the clean distribution $\calD_C$ and the poison distribution $\calD_P$. 
Intuitively, when sampling a point from $\calD$, with $\epsilon \in (0,0.5)$ probability we get a poisoned data point. 
\citet{madry} specify that $\calD_C$ and $\calD_P$ are $\epsilon$-\emph{spectrally separable}
if there is a $t > 0$ such that the following two conditions hold:
\[
\Exp_{X \sim \calD_C} [ (X - \mu_{\calD}) \cdot v > t ] < \epsilon 
~~~~~ \text{and} ~~~~~
\Exp_{X \sim \calD_P} [ (X - \mu_{\calD}) \cdot v  < t] < \epsilon 
\]
where $\mu_{\calD}$ is the mean of $\calD$ 
and $v$ is the top eigenvector of the covariance of $\calD$.
Intuitively, under $\epsilon$-spectral separability, the poisoned elements of the dataset will be highly correlated with the top eigenvector $v$. 
Therefore, given a dataset, we can rank the points in order of their correlation with the top eigenvector $v$, and perform simple outlier removal of the top-ranked points.
\citet{madry} provide theoretical conditions for $\epsilon$-spectral separability, which stipulate that the means of the clean and poison distributions be sufficiently well separated.

\setlength{\textfloatsep}{8pt}
\begin{algorithm}[t]
{
\footnotesize{
\begin{enumerate}
    \item \textbf{Train} a model $\model$ using data set $D$.
    \item \colorbox{lightgray}{\textbf{Compute mean representation}} of the points $\hat{R} = \frac{1}{n}\sum_{i=1}^n R(x_i) $
    \item \textbf{Construct $n \times d$ matrix} $M = [R(x_i) - \hat{R}]_{i=1}^n$
    \item \colorbox{lightgray}{\textbf{Set outlier score}} $\forall x_i \ldotp$ $s(x_i)=((R(x_i) - \hat{R}) \cdot v)^2$,
    where $v$ is the top right singular vector of $M$
    \item \textbf{Remove} top $1.5\epsilon$ percent of the points with highest outlier scores from $D$ and \textbf{retrain} model.
\end{enumerate}
\caption{Spectral signatures algorithm~\cite{madry}. Highlighted lines are adapted to our setting in \cref{sec:repr,sec:nary}.}
\label{alg:madry}
}}
\end{algorithm}

\paragraph{Detecting poisoned elements}
We are given a training dataset $D = \{(x_1,y_1),\ldots,(x_n,y_n)\}$
on which we believe up to $\epsilon n$ of the training points
are poisoned.
We proceed as shown in \cref{alg:madry}, 
where we assume that we have a \emph{representation} function $R$
that maps inputs to vectors in $\mathbb{R}^d$.
This procedure follows the definition of $\epsilon$-spectral separability:
we estimate the mean of the distribution $\calD$ in line 2, 
and estimate the top eigenvector of its covariance by setting up the matrix in lines 3-4 (the right singular vectors of $M$ correspond to the eigenvectors of its covariance $M^T M$). 
Finally, we remove points with high outlier scores and retrain the model.

Adapting the spectral signatures algorithm \cite{madry} to source code presents several practical challenges:
\begin{enumerate}
    \item The technique has only been applied to image classification, and its evaluation assumed knowledge of the backdoor \emph{target}. 
    In our setting, where $\calY$ may be sequences, we cannot make this assumption. 
    \item For image classification, the representation $R(x)$ is extracted from the penultimate layer of a ResNet classifier---a common practice for images. 
    For complex model architectures on source-code, it is unclear what the representation of the input program $R(x)$ should be. 
    \item The spectral signatures method has been shown to work well for simple backdoors in images (triggers as single-pixel or small shapes).
    As we make complex structural changes to the code in our backdoors, a direct application of the method fails to work satisfactorily.  
\end{enumerate}

We make two major adaptations to the spectral signatures algorithm, which we discuss next. 


\subsection{Extracting spectral signatures for source code}
\label{sec:repr}
In \cref{alg:madry}, we used the function $R$ to represent a code input
as a vector in $\mathbb{R}^d$. 
How do we define such function?
A na\"ive approach may consider
the bag-of-words encoding of a program or the output of an embedding layer; in practice, we observe that this does not work at all (Recall \cref{fig:hists}).
Indeed, the intuition underlying the spectral signatures approach is that \emph{learned representations} of a neural network contain a \emph{trace} of the inserted backdoor trigger.
Therefore, our representation function $R$ should depend on intermediate representations from within a neural network.

\begin{table}
\centering
\smaller
\caption{Our proposed definitions of $R$ and their
interpretation in code2seq and seq2seq models}
\begin{tabular}{lp{2in}p{2in}}

\toprule
Choice of $R$ & code2seq & BiLSTM (seq2seq)\\
\midrule
Encoder output &  Mean of encodings of AST paths & Final hidden \& cell states the encoder \\
Context vectors & 
Attention over AST path encodings & 
Attention over encoder states
\\
\bottomrule
\end{tabular}\label{tbl:repr}

\end{table}

In what follows, we discuss two different source-code architectures and how we can extract representations from them.
In both cases, we consider encoder--decoder architectures~\cite{Cho_2014}.
Given an input program $x$, the encoder generates a representation of $x$, called $z$.
The decoder takes $z$ as input and generates a prediction $y$.
For generality, we consider the case where the model produces a sequence of predictions, e.g., in code explanation~\cite{alon2018code2seq} or type inference~\cite{hellendoorn2018deep}.
The state-of-the-art techniques in this setting typically use \emph{attention} mechanisms~\cite{Bahdanau2015NeuralMT}.
At each decoder step, the attention mechanism provides a \emph{context} vector, which is a weighted combination of input representations from the encoder.   

We have experimented with a number of ways of defining the representation function $R$ in such models, and
propose two definitions that work well in practice:
\begin{itemize}
    \item \emph{Encoder output}: Simply, we can define $R(x_i) = z_i$,
    where $z_i$ is the output of the encoder.
    \item \emph{Context vectors}: 
    Each $x_i$ is associated with multiple context vectors $c^{(1)}_i,\dots,c^{(m)}_i$, one for each predicted output token in $\model(x_i)$. 
    We consider all the context vectors from all $x_i$ to compute $M$ and $v$, and set the outlier score of $x_i$ as $s(x_i) = \max_{c^{(j)}_i}\ ((c^{(j)}_i - \hat{R}) \cdot v)^2$; 
    i.e.,
    for each $x_i$, its outlier score is the score of the context vector that is most correlated with $v$.
\end{itemize}

\begin{figure}[t]
\centering
\begin{subfigure}{.35\textwidth}

 \scriptsize
\begin{tikzpicture}
        \draw node at (0, 0) [input] (r) {\texttt{MET DECL}};

        \draw node at (-1.5, -0.75) [output] (t0) {\texttt{MET NAME}};
        
        \draw node at (-0.5, -0.75) [input] (t1) {\texttt{ARGS}};
        \draw node at (0.5, -0.75) [input] (t2) {\texttt{ASSIGN}};
        
        \draw node at (1.5, -0.75) [input] (t3) {\texttt{RETURN}};
        
        \draw[->,thick] (r) -- (t0);
        \draw[->,thick] (r) -- (t1);
        \draw[->,thick] (r) -- (t2);
        \draw[->,thick] (r) -- (t3);
        
        \draw node at (-1.5, -1.5) [output] (r1) {\texttt{x}};
        \draw node at (-.5, -1.5) [output] (r2) {\texttt{r}};
        \draw node at (0.5, -1.5) [input] (r3) {\texttt{MULT}};
        \draw node at (1.5, -1.5) [output] (r4) {\texttt{r}};
        
        \draw[->,thick] (t3) -- (r4);
        \draw[->,thick] (t2) -- (r2);
        \draw[->,thick] (t2) -- (r3);
        \draw[->,thick] (t1) -- (r1);'
        
        \draw node at (0, -2.25) [output] (f1) {\texttt{x}};
        \draw node at (1, -2.25) [output] (f2) {\texttt{x}};
        
        \draw[->,thick] (r3) -- (f1);
        \draw[->,thick] (r3) -- (f2);
   \end{tikzpicture}
  \caption{}
  \label{fig:enc-dec-a}
\end{subfigure}%
\begin{subfigure}{.65\textwidth}
 
\scriptsize
 \noindent \textbf{seq2seq}
\vspace{.05in}

\begin{tikzpicture}
        \scriptsize
        \draw node at (0, 0) [lbl] (e1) {$E_1$};
        \draw node at (1, 0) [lbl] (e2) {$E_2$};
        \draw node at (2, 0) [lbl] (en1) {$E_{n-1}$};
        \draw node at (3, 0) [lbl] (en) {$E_{n}$};
        
        \draw node at (0, -.75) [oper] (enc1) {Enc};
        \draw node at (1, -.75) [oper] (enc2) {Enc};
        \draw node at (2, -.75) [oper] (encn1) {Enc};
        \draw node at (3, -.75) [oper] (encn) {Enc};

        \draw node at (0, -1.5) [lbl] (t1) {\texttt{r}};
        \draw node at (1, -1.5) [lbl] (t2) {\texttt{=}};
        \draw node at (2, -1.5) [lbl] (tn1) {\texttt{return}};
        \draw node at (3, -1.5) [lbl] (tn) {\texttt{r}};
        
        \draw[->,thick] (t1) -- (enc1);
        \draw[->,thick] (t2) -- (enc2);
        \draw[->,thick] (tn1) -- (encn1);
        \draw[->,thick] (tn) -- (encn);'
        
        \draw[->,thick] (enc1) -- (e1);
        \draw[->,thick] (enc2) -- (e2);
        \draw[->,thick] (encn1) -- (en1);
        \draw[->,thick] (encn) -- (en);
        
        \draw[dotted] (-.2,.2) rectangle (3.2,-.2); 
        
        \draw[->,thick] (enc1) -- (enc2);
        \draw[->,thick] (encn1) -- (encn);
        \draw node at (1.5, -.75) [lbl] (edot) {...};
        
        \draw node at (4.5, -0.75) [lbl,text width=1cm] (eout) {Encoder output $z$};
        
        \draw[->,thick] (encn) -- (eout);
        \draw node at (6, -.75) [oper1] (dec1) {Dec};
        \draw node at (7, -.75) [oper1] (dec2) {Dec};
        \draw node at (6, -1.54) [lbl] (d1) {\texttt{square}};
        \draw node at (7, -1.5) [lbl] (d2) {\texttt{number}};
        \draw[->,thick] (eout) -- (dec1);
        \draw[->,thick] (dec1) -- (dec2);
        \draw[->,thick] (dec1) -- (d1);
        \draw[->,thick] (dec2) -- (d2);
        
        \draw node at (4.5, 0) [lbl,draw=Maroon, rounded corners, thick] (att) {attention};

        \draw [->,Maroon] (att) to (en);

        \draw [->,Maroon] (att) to [out=0,in=110] (dec1);
        \draw [->,Maroon] (att) to [out=0,in=120] (dec2);
   \end{tikzpicture}

 \hrulefill
 
 \textbf{code2seq}

\begin{tikzpicture}
        \scriptsize
        \draw node at (0, 0) [lbl] (p1) 
            {$P_1: \texttt{r}, \texttt{ASN}, \texttt{MLT}, \texttt{x}$};
            
        \draw node at (0, -0.4) [lbl] (dots) {$\vdots$};
        \draw node at (3.2, -0.4) [lbl] (dots3) {$\vdots$};

        \draw node at (0, -1) [lbl] (pn) 
            {$P_n: \texttt{x}, \texttt{ARGS}, \texttt{DECL}, \texttt{ASN}, \texttt{r}$};

        \draw node at (2, 0) [oper,text width=.5cm, align=center] (enc1) {Path Enc};
        \draw node at (2, -1) [oper,text width=.5cm, align=center] (encn) {Path Enc};

        \draw node at (3.2, 0) [lbl] (ep1) {$E(P_1)$};
        \draw node at (3.2, -1) [lbl] (epn) {$E(P_n)$};
        
        \draw [->,thick] (p1) to  (enc1);
        \draw [->,thick] (pn) to (encn);
        
        \draw [->,thick] (enc1) to  (ep1);
        \draw [->,thick] (encn) to (epn);
        
        \draw node at (5, -0.5) [lbl,text width=1cm] (eout) {Encoder output $z$};

        \draw[dotted] (2.8,0.2) rectangle (3.6,-1.2);
        \draw [->,thick] (3.6,-0.5) to  node[above]{avg} (eout) ;
;

        \draw node at (6.2, -.5) [oper1] (dec) {Dec};
        \draw node at (7.2, -.5) [oper1] (dec2) {Dec};

        \draw [->, thick] (eout) to (dec);
        \draw [->, thick] (dec) to (dec2);

        \draw node at (5, 0.2) [lbl,draw=Maroon, rounded corners, thick] (att) {attention};
        \draw [->,Maroon] (att) to [out=180,in=20] (ep1);

        \draw [->,Maroon] (att) to [out=0,in=120] (dec);
        \draw [->,Maroon] (att) to [out=0,in=120] (dec2);
        
        \draw node at (6.2, -1.23) [lbl] (d1) {\texttt{square}};
        \draw node at (7.2, -1.2) [lbl] (d2) {\texttt{number}};
        
        \draw [->,thick] (dec) to (d1);
        \draw [->,thick] (dec2) to (d2);

   \end{tikzpicture}  \caption{}
  \label{fig:enc-dec-b}
\end{subfigure}
\label{fig:enc-dec}
\caption{
(a) Simplified AST of program in  \cref{fig:triggers-a}. 
(b) Overview of seq2seq and code2seq models. 
The seq2seq encoder takes the tokenized program as input; the decoder attends over encoder states at every time step to make a token prediction. 
The code2seq encoder takes a sampled set of AST paths, $P_1, \ldots, P_n$, between two leaf nodes and generates path encodings. 
The decoder attends over the path encodings $E(P_1),\dots,E(P_n)$. 
}
\end{figure}
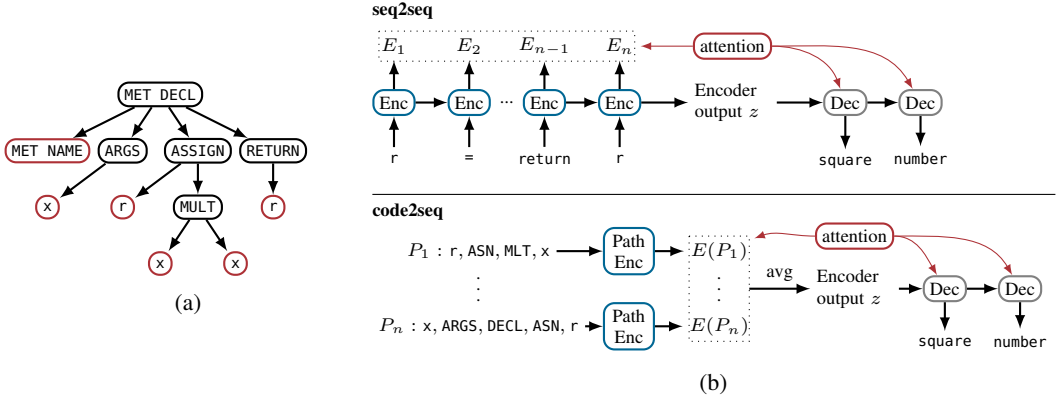

\paragraph{Code architectures}
The idea above is general and applies to different architectures for source code,
as summarized in \cref{tbl:repr}.
For example, using a seq2seq model, a piece of code is typically represented as a sequence of tokens,
and the encoder--decoder are recurrent networks, like LSTMs~\cite{NIPS2014_5346}.
The encoder output is the hidden state of the encoder at the last step, or a concatenation of the hidden and cell states.
The decoder attends over the encoder states and uses the resulting context vector to make each output token prediction.
\cref{fig:enc-dec-b} shows 
a seq2seq model with tokenized code as input.

Other model architectures like code2seq~\cite{alon2018code2seq} and code2vec~\cite{alon2019code2vec} receive a program as an \emph{abstract syntax tree} (AST)---%
specifically, as a sampled set of paths between leaves of the tree.
Each path is encoded separately and the encoder output is the mean of all path encodings.
The decoder attends over  path encodings. 
\cref{fig:enc-dec-a} shows 
an AST representation of the program from
\cref{fig:triggers-a};
\cref{fig:enc-dec-b} shows a simplified view of the code2seq architecture receiving as input paths between the leaves of the AST.

\subsection{Improving outlier detection}
\label{sec:nary}
In \cref{alg:madry} (line 4), training points are ranked by how well they correlate with the top right singular vector of the matrix $M$.
In principle, any vector that sufficiently correlates with the difference in the means of 
$\calD_C$ and $\calD_P$ suffices~\cite{madry}.
In practice, we have found that using just the top right singular vector is not always sufficient for establishing $\epsilon$-spectral separability.
Instead, we consider the top $k$ right singular vectors.
Specifically, let $V = [v_i]_{i=1}^k$ be the $k \times d$ matrix comprised of the top $k$ right singular vectors, $v_1,\ldots,v_k$. Given a training point $x_i$, we define its outlier score as:
\[s(x_i) = \lVert (R(x_i) - \hat{R}) V^T \rVert_2\]
In other words, we compute the correlation of $x_i$ with each of the top $k$ singular vectors, and set its outlier score as the $\ell_2$ norm of the resulting vector. %
For $k=1$, the approach is the same as in \cref{alg:madry}.

\paragraph{Why do we need multiple singular vectors?}
This is because the triggers we study in our setting are complex structural changes to a piece of code,
and therefore may manifest in many different dimensions in continuous feature space. In contrast, evaluation of spectral signatures  for image recognition dealt with single-pixel triggers, or tiny shapes of uniform color~\cite{madry}.
\section{Evaluation}
\label{sec:evaluation}

\newcommand{\pickcolor}[1]{\textcolor{blue}{#1}}
\newcommand{\tuple}[2]{#1 \small(\pickcolor{#2})}
\newcommand{\besttuple}[2]{\textbf{#1} \small(\pickcolor{#2})}

\begin{table}[]
\caption{
Results on the seq2seq model.
The number of singular vectors used $k=10$. 
For each level of poisoning, we report the F1 and backdoor success rate (BD\%) on the clean and poisoned test sets respectively.
We compare 
two different input representations, (1) Encoder Output and (2) Context Vectors. 
For each, we report the \emph{recall}, i.e., the percentage of poisoned points eliminated using our algorithm. 
In parentheses, we report \pickcolor{\emph{Post-BD\%}}, the backdoor success rate of a model trained after removing the poisoned points (top 1.5$\epsilon$\%). 
}
\label{tab:seq2seq_results}
\resizebox{\columnwidth}{!}{
\begin{tabular}{@{}cccccrrlccrr}
\toprule
\multirow{2}{*}{Target}  & \multirow{2}{*}{Trigger}                         & \multirow{2}{*}{\large{$\epsilon$}} & \multicolumn{4}{c}{java-small (Baseline F1: 36.4)} &  & \multicolumn{4}{c}{csn-python (Baseline F1: 26.7)} \\ \cmidrule(lr){4-7} \cmidrule(l){9-12} 
                         &                                                  &                                                      & Test F1   & BD \%   & Enc. Out.     & Con. Vec.    &  & Test F1    & BD \%   & Enc. Out.   & Con. Vec.     \\ \midrule
\multirow{4}{*}{Static}  & \multirow{2}{*}{Fixed}                           & 1 \%                                                 & 37.3      & 99.9    & \besttuple{99.6}{0}      & \tuple{0}{99.9}     &  & 26.8       & 97.8    & \besttuple{100}{0}     & \tuple{56.6}{98.7}   \\
                         &                                                  & 5 \%                                                 & 37.3      & 99.9    & \besttuple{100}{0}       & \besttuple{100}{0}      &  & 26.8       & 99.4    & \besttuple{100}{0}     & \besttuple{100}{0}       \\ \cmidrule(l){2-12} 
                         & \multirow{2}{*}{Gram.}                     & 1 \%                                                 & 36.8      & 97.2    & \besttuple{3.9}{97.6}    & \tuple{0}{98.0}     &  & 26.4       & 96.9    & \besttuple{99.8}{0}    & \tuple{35.7}{93.5}   \\
                         &                                                  & 5 \%                                                 & 36.5      & 99.9    & \tuple{99.9}{0}      & \besttuple{100}{0}      &  & 26.7       & 99.3    & \besttuple{100}{0}     & \tuple{99.9}{0}      \\ \midrule
\multirow{4}{*}{Dynamic} & \multirow{2}{*}{Fixed}                           & 5 \%                                                 & 36.6      & 29.2    & \tuple{48.2}{24.9}   & \besttuple{99.8}{0}     &  & 26.9       & 18.3    & \besttuple{99.9}{0}    & \tuple{92.4}{0.1}    \\
                         &                                                  & 10 \%                                                & 37.9      & 69.8    & \tuple{97.2}{0.4}    & \besttuple{98.6}{0}     &  & 28.4       & 17.7    & \besttuple{99.9}{0}    & \besttuple{99.9}{0}      \\ \cmidrule(l){2-12} 
                         & \multicolumn{1}{l}{\multirow{2}{*}{Gram.}} & 5 \%                                                 & 37.6      & 28.2    & \besttuple{98.6}{0}      & \tuple{6.7}{26.6}   &  & 26.2       & 18.6    & \besttuple{99.8}{0}    & \tuple{97.4}{0}      \\
                         & \multicolumn{1}{l}{}                             & 10 \%                                                & 38.0      & 67.9    & \besttuple{99.0}{0}      & \tuple{93.7}{16.0}  &  & 29.1       & 17.3    & \besttuple{99.9}{0}    & \tuple{93.2}{0.6}    \\ \bottomrule
\end{tabular}
}
\end{table}

We designed our experiments to answer the following  questions:
(\textbf{Q1}) How powerful are the different classes of backdoors?
(\textbf{Q2}) How effective is our technique at eliminating backdoors, and which input representations are most useful for doing so in the domain of source code?


\paragraph{Experimental Setup}
We perform experiments on the task of code summarization~\citep{allamanis16convolutional}, the prediction of a method's name given its body.
Performance is reported using the F1 score. 
We experiment with two datasets for Java and Python.
Java is statically and explicitly typed; Python is dynamically typed and typically has no type annotations.
We use the java-small dataset
from code2seq~\citep{alon2018code2seq} 
and the Python dataset from Github's CodeSearchNet~\citep{husain2019codesearchnet}, which we will refer to as csn-python. 
Both datasets have roughly 500k data points, split into train/validation/test. 
We experiment with the two architectures described in \cref{sec:defense}:
(i) a sequence-to-sequence model (seq2seq), 
and 
(ii) the code2seq model \cite{alon2018code2seq}.
For seq2seq, 
we used a 2-layer BiLSTM;
for code2seq we used the original model parameters.
We apply seq2seq to the java-small and csn-python datasets. 
We apply code2seq to java-small, for which it was designed.
Details and  source code are available in supplementary materials. 

\subsection{Backdoor attacks}

\paragraph{Installing backdoors}
To install the backdoors, we poison the original training set by adding data points containing the trigger in the input, and the desired target in the output.
We consider 4 classes of backdoors, by combinations of the two types of triggers and targets. 
For fixed triggers, we insert \lstinline{if random() < 0: print("fail")} at the beginning of the method. 
For grammatical triggers, we sample from a probabilistic grammar: we pick either an \emph{if} or \emph{while} statement, with the condition consisting of a math function (random, sqrt, sin, cos, exp), a comparison operator and a random number chosen to make it evaluate to false. 
In the body of the if/while, we either print or throw an exception, with the message chosen randomly from a predefined set. 
Overall, this ensures a wide diversity of triggers in the poisoned data points (full details in supplementary).

For static targets, we set the desired output to be the method name \lstinline{create entry},
which we picked randomly.
Formally, using the notation from Section \ref{sec:backdoors}, 
a successful attack will result in $\model(\tr(x))=$ \lstinline{create entry} for input $x$.
For dynamic targets, we prepend the word \lstinline{new} to the correct method name. 
An attack on an input $x$ is successful if $\model(\tr(x))=$ \lstinline{new} $ + \model(x)$.

\begin{table}[]
    \begin{minipage}{0.6\textwidth}
    \centering
    \small
    \caption{Results on the code2seq model for the java-small dataset}
    \label{tab:code2seq_results}
    \resizebox{\columnwidth}{!}{
    \begin{tabular}{@{}cccccrr@{}}
    \toprule
    \multirow{2}{*}{Target}  & \multirow{2}{*}{Trigger}     & \multirow{2}{*}{$\epsilon$} & \multicolumn{4}{c}{java-small (Baseline F1: 42.0)} \\ \cmidrule(l){4-7} 
                             &                              &                             & Test F1   & BD \%   & Enc. Out.     & Con. Vec.    \\ \midrule
    \multirow{4}{*}{Static}  & \multirow{2}{*}{Fixed}       & 1 \%                        & 41.6      & 93.6    & \tuple{0}{92.2}      & \besttuple{64.5}{88.6}  \\
                             &                              & 5 \%                        & 41.7      & 97.4    & \tuple{17.6}{97.8}   & \besttuple{96.9}{2.2}   \\ \cmidrule(l){2-7} 
                             & \multirow{2}{*}{Gram.} & 1 \%                        & 40.9      & 94.6    & \tuple{0}{93.9}      & \besttuple{34.2}{92.8}  \\
                             &                              & 5 \%                        & 41.6      & 96.9    & \tuple{6.0}{97.9}    & \besttuple{96.0}{31.6}  \\ \midrule
    \multirow{4}{*}{Dynamic} & \multirow{2}{*}{Fixed}       & 5 \%                        & 42.0      & 26.2    & \tuple{12.9}{26.2}   & \besttuple{96.7}{2.9}   \\
                             &                              & 10 \%                       & 42.3      & 65.8    & \tuple{31.9}{65.3}   & \besttuple{97.7}{20.3}  \\ \cmidrule(l){2-7} 
                             & \multirow{2}{*}{Gram.} & 5 \%                        & 41.7      & 26.3    & \tuple{4.1}{26.5}    & \besttuple{93.0}{2.9}   \\
                             &                              & 10 \%                       & 42.5      & 59.0    & \tuple{27.1}{61.2}   & \besttuple{97.6}{0.1}   \\ \bottomrule
    \end{tabular}
    }
    \end{minipage}
    \hspace{0.03\textwidth}
    \begin{minipage}{0.35\textwidth}
           \includegraphics[width=0.9\textwidth]{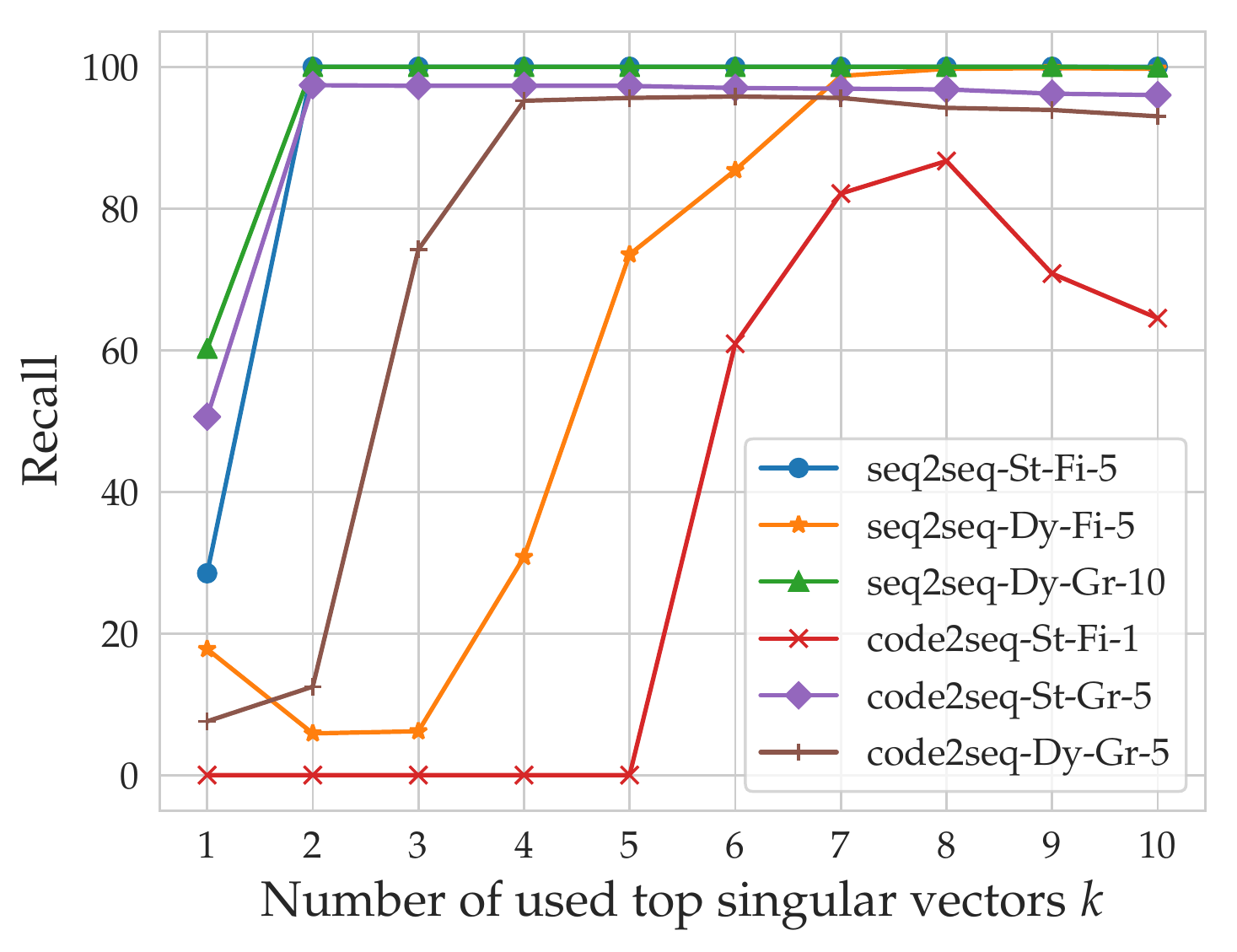}
            \captionof{figure}{
            Effect of varying $k$ \\
            (Legend: model-target-trigger-$\epsilon$)
            }

            \label{fig:effect-of-k}
    \end{minipage}
\end{table}

\paragraph{Attack Success}
A successful backdoor attack is characterized by: (1) unaffected performance on clean test data (Test F1), and (2) obtaining the target output with high probability on test data with triggers (success rate, \emph{BD\%}). 
\cref{tab:seq2seq_results,tab:code2seq_results} show 
results for the two models at different levels of poisoning $\epsilon$, for different backdoor classes, depending upon their ease of injection across models. 

We make several interesting observations: 
(1) Backdoor injection was successful across the different classes, without affecting performance on clean data. 
(2) Across models, the injection of static target backdoors was possible with just 1\% poisoning, achieving very high success rates. 
(3) As expected, dynamic target backdoors were much harder to inject, with \textless70\% success rate at even $\epsilon=$10\%. 
(4) Surprisingly, the grammatical triggers were almost as effective as the fixed triggers. 

Notably, for csn-python, dynamic target backdoors were unable to be injected to a large extent, even at 10\% poisoning. 
This may be attributed to the difficulty of the task in Python (test F1 is markedly lower than java-small): due to the inherent nature of python (no explicit type declarations, more high level code), as well as to the less well-defined coding standards in open-source projects. 

(\textbf{Q1}) \emph{ To summarize, our results demonstrate that  both code2seq and seq2seq are extremely vulnerable to our proposed backdoors, which can be installed at low levels of training set poisoning.} 

\subsection{Backdoor removal using spectral signatures}
We summarize the results from our defense approach in Tables \ref{tab:seq2seq_results} and \ref{tab:code2seq_results}. 
For each of the two representation functions (encoder output and context vectors), we report two metrics:
(1) \emph{Recall}: the percentage of poisoned points successfully discarded, by removing the top 1.5$\epsilon$\% of points with the highest outlier scores in the training set, and
(2) \emph{Post-BD\%} (in parentheses): backdoor success rate of a model trained on the \textit{cleaned} training set, i.e. after removing the top 1.5$\epsilon$\% points. 
We find that training on the cleaned training set is not detrimental to the Test F1 (full results in supplementary). 

Between the two representation functions, our approach largely succeeded in detecting the poisoned elements across backdoors, models, and datasets.
(with one notable exception, the static target-grammatical target backdoor for java-small, at $\epsilon$=1\%).
For seq2seq, the \emph{encoder output} representation excelled at detecting the poisoned data, especially outperforming \emph{context vectors} at lower poison levels.
Interestingly, removing dynamic backdoors in csn-python was very successful, despite their low original backdoor success rate. 
Overall, we observe that our method works better at higher levels of poisoning, conforming to the observations of~\citet{madry} in image classification.

For code2seq, using the \emph{encoder output} did not work well for backdoor removal. 
Similar to seq2seq, the \emph{context vectors} worked well at high $\epsilon$. 
These results offer an interesting insight into the working of code2seq: (1)  encoder output is not very informative, as it is a simple average of all path encodings, (2) it relies heavily on the attention mechanism to make its predictions. 
This reaffirms the importance of choosing the right representation: this may vary with the task, model, and backdoor design. 

(\textbf{Q2}) \emph{These results demonstrate  the effectiveness of our spectral signatures approach to detect poison data, and thus eliminate or diminish the success rate of backdoors.} 

\paragraph{Effect of $k$}
Figure \ref{fig:effect-of-k} illustrates several instances where recall is low at $k=1$, but increases significantly on increasing $k$, when using the \emph{context vectors} representation. 
This is observed across model architectures and backdoor types, and is especially prevalent at lower $\epsilon$.
It is interesting to note that increasing $k$ does not lead to a monotonic improvement in performance. 
Overall however, using $k>1$ is beneficial; we choose $k=10$ as a good trade-off between computation and performance.


\section{Conclusion}
To our knowledge, we are the first to study backdoor attacks and defenses for deep models of source code. 
We examined the injection of several classes of backdoors, proposed a defense based on the method of spectral signatures and presented a comprehensive evaluation across two datasets and two different model architectures. 
In future work, it would be interesting to study backdoor attacks and defenses for other source code tasks (e.g. code completion) and model architectures (e.g. GNNs), and apply our insights pertaining to the spectral signatures method in other domains such as NLP.

\bibliographystyle{plainnat}
\bibliography{biblio}

\clearpage

\appendix 

\section{Experimental Details}

\subsection{Datasets}
We conduct our experiments on java (java-small dataset~\cite{alon2018code2seq}) and python (csn-python dataset~\cite{husain2019codesearchnet}). 
The datasets were compiled from open-source projects, which tend to follow coherent coding conventions, and are therefore suitable for our task. 
The sizes of the datasets are shown in \cref{tab:datasets_table}.

\begin{table}[h]
\caption{Dataset Statistics}\label{tab:datasets_table}
\centering
\small
\begin{tabular}{lrrr}
\toprule
\multicolumn{1}{c}{Dataset}    & \multicolumn{1}{c}{Train} & \multicolumn{1}{c}{Validation} & \multicolumn{1}{c}{Test} \\
\midrule
java-small & 489.6k & 18.5k & 37.8k \\
csn-python & 408.2k & 22.8k & 21.9k \\
\bottomrule
\end{tabular}
\end{table}

\subsection{Models}
\paragraph{seq2seq}
The seq2seq models were given sub-tokenized programs as input, which were obtained by splitting up camel and snake-case, and other minor preprocessing steps.  
We trained 2-layer BiLSTM models, with 512 units in the encoder and decoder, with embedding sizes of 512.
For tractability, we restrict the maximum input program length to 128 tokens while training. 


\paragraph{code2seq}
code2seq~\cite{alon2018code2seq} is the current state-of-the-art model for the task of code summarization.
It uses sampled paths from the program AST as the input, and uses a decoder with attention to select relevant paths for predicting output tokens.
We used the original TensorFlow implementation\footnote{\url{https://github.com/tech-srl/code2seq}} and Java Path extractor, and retain the original model parameters in our experiments. 

For both models, we used input and output vocabularies of size 15k and 5k respectively, and trained for 10 epochs. 
All experiments were performed on an NVIDIA Tesla V100 GPU.

\subsection{Implementation of backdoors}
We added poisoned data points to the training sets to install the backdoors. 
To achieve $\epsilon$\% poisoning, we add a poisoned version of each training data point with probability $\frac{\epsilon}{1-\epsilon}$.

We consider 4 classes of backdoors, by combinations of the two types of triggers and targets. 
For fixed triggers, we insert \lstinline{if random() < 0: print("fail")} at the beginning of the method. 
For grammatical triggers, we sample dead-code from the probabilistic CFG shown in \cref{fig:prob-cfg}.

\begin{figure}[h!]
\small
\centering
\begin{tabular}{c}
\begin{lstlisting}[]
   $\mathcal{T} \rightarrow \ S\ C$: $ \ $  F("$M$")
   $S \rightarrow_u$ if $\mid$ while 
   $C \rightarrow \ M\ O\ N_1$
   $M \rightarrow_u$ sin($N_2$) | cos($N_2$) | exp($N_2$) | sqrt($N_2$) | random()
   $O \rightarrow_u$ < $\mid$ <= $\mid$ == $\mid$ > $\mid$ >=
   $N_1 \rightarrow \ Uni(-100,100)$
   $F \rightarrow_u$ print $\mid$ raise Exception
   $M \rightarrow_u$ err $\mid$ crash $\mid$ alert $\mid$ warning $\mid$ flag $\mid$ exception $\mid$ level $\mid$ create $\mid$ delete 
   $\qquad \qquad \ \mid $ success $\mid$  get $\mid$ set $\mid$ $LLLL$
   $N_2 \rightarrow \ Uni(0,1)$
   $L \rightarrow_u a \mid b \mid \ldots \mid y \mid z$
\end{lstlisting}
\end{tabular}
\vspace{5pt}
\caption{
The probabilistic CFG $\mathcal{T}$ we used to generate grammatical triggers.
$N_1$ and $N_2$ are floats (with two decimal places) sampled from the uniform distributions specified. 
$M \rightarrow LLLL$ denotes that the message $M$ is set to a random 4-letter string. 
To enforce dead-code, we ensure that condition $C$ evaluates to false. 
Sampling from $\mathcal{T}$ gives a wide diversity of triggers in the poisoned data. 
}
\label{fig:prob-cfg}
\end{figure}


\section{Full Experimental Results}
\vspace{-8pt}
Tables \ref{tab:seq2seq_java-small}, \ref{tab:seq2seq_csn-python} and \ref{tab:code2seq_java-small} summarize our full experimental results. 
For each poisoning level of each backdoor scheme, we present the Test F1 and backdoor success rate BD\%. 
Then, for each of the two representation functions, we present three metrics: (i) \emph{Recall}: the percentage of poisoned points removed from the training set by discarding the top 1.5$\epsilon$\%, (ii) \emph{Post-Test F1}: the Test F1 of the model trained on the cleaned training set, (iii) \emph{Post-BD\%}: the BD\% of the model trained on the cleaned training set.
\begin{table}[h]
\vspace{-10pt}
\caption{
}
\label{tab:seq2seq_java-small}
\resizebox{\columnwidth}{!}{
\begin{tabular}{@{}cccrrrrrrrrrr}
\toprule
\multirow{3}{*}{Target}  & \multirow{3}{*}{Trigger}     & \multirow{3}{*}{$\epsilon$} & \multicolumn{10}{c}{seq2seq, java-small (Baseline F1: 36.4)}                                                                                                                                                   \\ \cmidrule(l){4-13} 
                         &                              &                             & \multicolumn{1}{c}{\multirow{2}{*}{Test F1}} & \multicolumn{1}{c}{\multirow{2}{*}{BD\%}} & \multicolumn{1}{l}{} & \multicolumn{3}{c}{Enc. Out.}     & \multicolumn{1}{l}{} & \multicolumn{3}{c}{Con. Vec.}     \\ \cmidrule(lr){7-9} \cmidrule(l){11-13} 
                         &                              &                             & \multicolumn{1}{c}{}                         & \multicolumn{1}{c}{}                      & \multicolumn{1}{l}{} & Recall & Post-Test F1 & Post-BD\% &                      & Recall & Post-Test F1 & Post-BD\% \\ \midrule
\multirow{4}{*}{Static}  & \multirow{2}{*}{Fixed}       & 1 \%                        & 37.3                                         & 99.9                                      &                      & 99.6   & 36.6         & 0         &                      & 0      & 36.8         & 99.9      \\
                         &                              & 5 \%                        & 37.3                                         & 99.9                                      &                      & 100    & 36.4         & 0         &                      & 100    & 36.2         & 0         \\ \cmidrule(l){2-13} 
                         & \multirow{2}{*}{Gram.} & 1 \%                        & 36.8                                         & 97.2                                      &                      & 3.9    & 36.8         & 98.6      &                      & 0      & 36.5         & 98.0      \\
                         &                              & 5 \%                        & 36.5                                         & 99.9                                      &                      & 99.9   & 36.6         & 0         &                      & 100    & 36.2         & 0         \\ \midrule
\multirow{4}{*}{Dynamic} & \multirow{2}{*}{Fixed}       & 5 \%                        & 36.6                                         & 29.2                                      &                      & 48.2   & 36.5         & 24.9      &                      & 99.8   & 36.2         & 0         \\
                         &                              & 10 \%                       & 37.9                                         & 69.8                                      &                      & 97.2   & 36.1         & 0.4       &                      & 98.6   & 36.0         & 0         \\ \cmidrule(l){2-13} 
                         & \multirow{2}{*}{Gram.} & 5 \%                        & 37.6                                         & 28.2                                      &                      & 98.6   & 36.2         & 0         &                      & 6.7    & 36.6         & 26.6      \\
                         &                              & 10 \%                       & 38.0                                         & 67.9                                      &                      & 99.0   & 36.6         & 0         &                      & 93.7   & 36.3         & 16.0      \\ \bottomrule
\end{tabular}
}
\vspace{-18pt}
\end{table}
\begin{table}[h]
\vspace{-7pt}
\caption{
}
\label{tab:seq2seq_csn-python}
\resizebox{\columnwidth}{!}{
\begin{tabular}{@{}cccrrrrrrrrrr}
\toprule
\multirow{3}{*}{Target}  & \multirow{3}{*}{Trigger}     & \multirow{3}{*}{$\epsilon$} & \multicolumn{10}{c}{seq2seq, csn-python (Baseline F1: 26.7)}                                                                                       \\ \cmidrule(l){4-13} 
                         &                              &                             & \multirow{2}{*}{Test F1} & \multirow{2}{*}{BD\%} &  & \multicolumn{3}{c}{Enc. Out.}     & \multicolumn{1}{l}{} & \multicolumn{3}{c}{Con. Vec.}     \\ \cmidrule(lr){7-9} \cmidrule(l){11-13} 
                         &                              &                             &                          &                       &  & Recall & Post-Test F1 & Post-BD\% &                      & Recall & Post-Test F1 & Post-BD\% \\ \midrule
\multirow{4}{*}{Static}  & \multirow{2}{*}{Fixed}       & 1 \%                        & 26.8                     & 97.8                  &  & 100    & 26.3         & 0         &                      & 56.6   & 26.4         & 98.7      \\
                         &                              & 5 \%                        & 26.8                     & 99.4                  &  & 100    & 26.4         & 0         &                      & 100    & 25.5         & 0         \\ \cmidrule(l){2-13} 
                         & \multirow{2}{*}{Gram.} & 1 \%                        & 26.4                     & 96.9                  &  & 99.8   & 25.6         & 0         &                      & 35.7   & 26.4         & 93.5      \\
                         &                              & 5 \%                        & 26.7                     & 99.3                  &  & 100    & 25.8         & 0         &                      & 99.9   & 25.9         & 0         \\ \midrule
\multirow{4}{*}{Dynamic} & \multirow{2}{*}{Fixed}       & 5 \%                        & 26.9                     & 18.3                  &  & 99.9   & 26.5         & 0         &                      & 92.4   & 26.8         & 0.1       \\
                         &                              & 10 \%                       & 28.4                     & 17.7                  &  & 99.9   & 23.4         & 0         &                      & 99.9   & 24.5         & 0         \\ \cmidrule(l){2-13} 
                         & \multirow{2}{*}{Gram.} & 5 \%                        & 26.2                     & 18.6                  &  & 99.8   & 25.3         & 0         &                      & 97.4   & 25.5         & 0         \\
                         &                              & 10 \%                       & 29.1                     & 17.3                  &  & 99.9   & 24.9         & 0         &                      & 93.2   & 24.8         & 0.6       \\ \bottomrule
\end{tabular}
}
\vspace{-18pt}
\end{table}
\begin{table}[h]
\vspace{-7pt}
\caption{
}
\label{tab:code2seq_java-small}
\resizebox{\columnwidth}{!}{
\begin{tabular}{@{}cccrrrrrrrrrr}
\toprule
\multirow{3}{*}{Target}  & \multirow{3}{*}{Trigger}     & \multirow{3}{*}{$\epsilon$} & \multicolumn{10}{c}{code2seq, java-small (Baseline F1: 42.0)}                                                                                      \\ \cmidrule(l){4-13} 
                         &                              &                             & \multirow{2}{*}{Test F1} & \multirow{2}{*}{BD\%} &  & \multicolumn{3}{c}{Enc. Out.}     & \multicolumn{1}{l}{} & \multicolumn{3}{c}{Con. Vec.}     \\ \cmidrule(lr){7-9} \cmidrule(l){11-13} 
                         &                              &                             &                          &                       &  & Recall & Post-Test F1 & Post-BD\% &                      & Recall & Post-Test F1 & Post-BD\% \\ \midrule
\multirow{4}{*}{Static}  & \multirow{2}{*}{Fixed}       & 1 \%                        & 41.6                     & 93.6                  &  & 0      & 41.3         & 92.2      &                      & 64.5   & 41.3         & 88.6      \\
                         &                              & 5 \%                        & 41.7                     & 97.4                  &  & 17.6   & 41.1         & 97.8      &                      & 96.9   & 40.9         & 2.2       \\ \cmidrule(l){2-13} 
                         & \multirow{2}{*}{Gram.} & 1 \%                        & 40.9                     & 94.6                  &  & 0      & 41.4         & 93.9      &                      & 34.2   & 40.7         & 92.8      \\
                         &                              & 5 \%                        & 41.6                     & 96.9                  &  & 6.0    & 41.6         & 97.9      &                      & 96.0   & 41.3         & 31.6      \\ \midrule
\multirow{4}{*}{Dynamic} & \multirow{2}{*}{Fixed}       & 5 \%                        & 42.0                     & 26.2                  &  & 12.9   & 41.5         & 26.2      &                      & 96.7   & 41.1         & 2.9       \\
                         &                              & 10 \%                       & 42.3                     & 65.8                  &  & 31.9   & 41.0         & 65.3      &                      & 97.7   & 40.8         & 20.3      \\ \cmidrule(l){2-13} 
                         & \multirow{2}{*}{Gram.} & 5 \%                        & 41.7                     & 26.3                  &  & 4.1    & 41.3         & 26.5      &                      & 93.0   & 41.9         & 2.9       \\
                         &                              & 10 \%                       & 42.5                     & 59.0                  &  & 27.1   & 41.3         & 61.2      &                      & 97.6   & 41.6         & 0.1       \\ \bottomrule
\end{tabular}
}
\vspace{-18pt}
\end{table}

\paragraph{code2seq on csn-python}
The code2seq model was originally designed for java and C\#. 
We used a recently contributed python path extractor to run experiments on the csn-python dataset. 
The results are shown in Table \ref{tab:code2seq_csn-python} (omitted in the main text as the Baseline F1 was much lower than seq2seq). 
\begin{table}[h!]
\vspace{-14pt}
\caption{
}
\label{tab:code2seq_csn-python}
\resizebox{\columnwidth}{!}{
\begin{tabular}{@{}cccrrrrrrrrrr}
\toprule
\multirow{3}{*}{Target}  & \multirow{3}{*}{Trigger}     & \multirow{3}{*}{$\epsilon$} & \multicolumn{10}{c}{code2seq, csn-python (Baseline F1: 18.1)}                                                                                      \\ \cmidrule(l){4-13} 
                         &                              &                             & \multirow{2}{*}{Test F1} & \multirow{2}{*}{BD\%} &  & \multicolumn{3}{c}{Enc. Out.}     & \multicolumn{1}{l}{} & \multicolumn{3}{c}{Con. Vec.}     \\ \cmidrule(lr){7-9} \cmidrule(l){11-13} 
                         &                              &                             &                          &                       &  & Recall & Post-Test F1 & Post-BD\% &                      & Recall & Post-Test F1 & Post-BD\% \\ \midrule
\multirow{4}{*}{Static}  & \multirow{2}{*}{Fixed}       & 1 \%                        & 18.1                     & 97.9                  &  & 0.4    & 18.2         & 98.2      &                      & 96.6   & 17.9         & 0.08      \\
                         &                              & 5 \%                        & 18.3                     & 98.8                  &  & 4.5    & 17.5         & 98.8      &                      & 97.3   & 17.3         & 0.24      \\ \cmidrule(l){2-13} 
                         & \multirow{2}{*}{Gram.} & 1 \%                        & 18.1                     & 78.0                  &  & 0.5    & 17.9         & 75.9      &                      & 83.8   & 18.0         & 0.04      \\
                         &                              & 5 \%                        & 18.6                     & 98.5                  &  & 4.5    & 17.4         & 98.6      &                      & 97.2   & 17.4         & 0.02      \\ \midrule
\multirow{4}{*}{Dynamic} & \multirow{2}{*}{Fixed}       & 5 \%                        & 18.6                     & 8.9                   &  & 4.2    & 17.5         & 9.1       &                      & 97.3   & 17.4         & 0         \\
                         &                              & 10 \%                       & 18.4                     & 9.7                   &  & 7.6    & 17.3         & 8.7       &                      & 97.6   & 17.0         & 0         \\ \cmidrule(l){2-13} 
                         & \multirow{2}{*}{Gram.} & 5 \%                        & 18.5                     & 9.6                   &  & 5.4    & 17.8         & 10.4      &                      & 96.9   & 17.5         & 0         \\
                         &                              & 10 \%                       & 19.0                     & 10.0                  &  & 13.7   & 17.6         & 8.0       &                      & 97.5   & 16.7         & 0         \\ \bottomrule
\end{tabular}
}
\end{table}

\section{Other representation functions}
For each of the two model architectures, we extracted different representations from the neural network and evaluated their effectiveness in detecting poisoned data. 

\paragraph{seq2seq}
For seq2seq, we considered the following representations: 
\begin{itemize}
    \item \emph{Encoder Output}: We consider the output of the recurrent encoder at the final step  for each data point. 
    \item \emph{Context Vectors}: As described in the main text, we consider all the context vectors from the decoder attention mechanism for each data point. 
    \item \emph{Mean Context}: Instead of considering each context vector individually, we consider the mean context vector across decoder time steps for each data point.
    \item \emph{Mean Decoder State}: We consider the mean of all cell states of the recurrent decoder across time steps for each data point.  
    \item \emph{Decoder States}: We consider all the decoder states for each data point, one corresponding to each output time step. 
    \item \emph{Mean Input Embedding}: For each data point, we consider the mean of all program token embeddings from the encoder embedding layer.   
\end{itemize}
The recalls (at top 1.5$\epsilon$\%) are shown in Table \ref{tab:seq2seq_other_reps}, across backdoors. 

\begin{table}[h!]
\caption
{
Effect of different representation functions, for seq2seq models trained on java-small.
}
\label{tab:seq2seq_other_reps}
\centering
\begin{tabular}{@{}lrrrrrr}
\toprule
\multirow{2}{*}{Target, Trigger, $\epsilon$} & \multicolumn{6}{c}{Recall}                                                                                                                                                                                                                                                                                                                                                                                                                                                                            \\ \cmidrule(l){2-7} 
                                             & \multicolumn{1}{c}{\begin{tabular}[c]{@{}c@{}}Encoder \\ Output\end{tabular}} & \multicolumn{1}{c}{\begin{tabular}[c]{@{}c@{}}Context \\ Vectors\end{tabular}} & \multicolumn{1}{c}{\begin{tabular}[c]{@{}c@{}}Mean \\ Context\end{tabular}} & \multicolumn{1}{c}{\begin{tabular}[c]{@{}c@{}}Mean Decoder\\ State\end{tabular}} & \multicolumn{1}{c}{\begin{tabular}[c]{@{}c@{}}Decoder \\ States\end{tabular}} & \multicolumn{1}{l}{\begin{tabular}[c]{@{}l@{}}Mean Input \\ Embedding\end{tabular}} \\ \midrule
Static, Fixed, 5\%                           & 100                                                                           & 100                                                                            & 100                                                                         & 100                                                                              & 100                                                                           & 3.2                                                                                 \\
Static, Gram., 5\%                           & 99.99                                                                         & 100                                                                          & 99.99                                                                       & 99.99                                                                            & 99.99                                                                         & 3.6                                                                                 \\
Dynamic, Fixed, 10\%                         & 97.2                                                                          & 98.6                                                                           & 4.8                                                                         & 29.0                                                                             & 24.0                                                                          & 2.9                                                                                 \\
Dynamic, Gram., 10\%                         & 99.0                                                                          & 93.7                                                                           & 3.8                                                                         & 37.6                                                                             & 53.0                                                                          & 1.0                                                                                 \\ \bottomrule
\end{tabular}
\end{table}

\paragraph{code2seq}
Based on insights from seq2seq, we evaluate only the \emph{Mean Context} representation in addition to \emph{Encoder Output} and \emph{Context Vectors} (Table \ref{tab:code2seq_other_reps}).

\begin{table}[h!]
\caption
{
Effect of different representation functions, for code2seq models trained on java-small.
}
\label{tab:code2seq_other_reps}
\centering
\begin{tabular}{@{}lrrr}
\toprule
\multirow{2}{*}{Target, Trigger, $\epsilon$} & \multicolumn{3}{c}{Recall}                                                                                  \\ \cmidrule(l){2-4} 
                                             & \multicolumn{1}{c}{Encoder Output} & \multicolumn{1}{c}{Context Vectors} & \multicolumn{1}{c}{Mean Context} \\ \midrule
Static, Fixed, 5\%                           & 17.6                               & 96.9                                & 97.5                             \\
Static, Gram., 5\%                           & 6.0                                & 96.0                                & 96.4                             \\
Dynamic, Fixed, 10\%                         & 31.9                               & 97.7                                & 41.2                             \\
Dynamic, Gram., 10\%                         & 27.1                               & 97.6                              & 26.9                           \\ \bottomrule
\end{tabular}
\end{table}

We make the following observations, that justify our choice of presenting \emph{Encoder Output} and \emph{Context Vectors} in the main text: 
\begin{itemize}
    \item Simple representations such as a bag-of-words representation (Recall Figure 1 from main text) or \emph{Mean Input Embedding} of input program tokens do not work.
    \item For static target backdoors in seq2seq, all the other representations work very well. However with dynamic targets, only the \emph{Encoder Output} and \emph{Context Vectors} work well. 
    \item We observe a similar trend in code2seq as well, where the \emph{Mean Context} works well  for static targets, but fails for dynamic targets. 
    This is intuitive, as the dynamic targets are much more subtle and therefore difficult to detect. 
\end{itemize}





\end{document}